\documentclass{article}
\usepackage{spconf,amsmath,amssymb,graphicx}
\usepackage{adjustbox}
\usepackage[table,xcdraw]{xcolor}
\usepackage{multirow}

\title{Towards ASR Robust Spoken Language Understanding Through In-Context Learning With Word Confusion Networks}
\name{
\begin{tabular}{@{}c@{}}
Kevin Everson$^{1^*,2}$\thanks{$^*$Work done as an applied scientist intern.}, Yile Gu$^{1}$, Huck Yang$^{1}$, Prashanth Gurunath Shivakumar$^{1}$, Guan-Ting Lin$^{1^*,3}$,\\ Jari Kolehmainen$^{1}$, Ivan Bulyko$^{1}$, Ankur Gandhe$^{1}$, Shalini Ghosh$^{1}$, Wael Hamza$^{1}$,\\ Hung-yi Lee$^{3}$, Ariya Rastrow$^{1}$, Andreas Stolcke$^{1}$
\end{tabular}}
\address{$^{1}$Amazon Alexa AI, USA, $^{2}$University of Washington, USA, $^{3}$National Taiwan University, Taiwan}
\begin{document}
\ninept
\maketitle
\begin{abstract}
In the realm of spoken language understanding (SLU), numerous natural language understanding (NLU) methodologies have been adapted by supplying large language models (LLMs) with transcribed speech instead of conventional written text. In real-world scenarios, prior to input into an LLM, an automated speech recognition (ASR) system generates an output transcript hypothesis, where inherent errors can degrade subsequent SLU tasks. Here we introduce a method that utilizes the ASR system's lattice output instead of relying solely on the top hypothesis, aiming to encapsulate speech ambiguities and enhance SLU outcomes. Our in-context learning experiments, covering spoken question answering and intent classification, underline the LLM's resilience to noisy speech transcripts with the help of word confusion networks from lattices, bridging the SLU performance gap between using the top ASR hypothesis and an oracle upper bound. Additionally, we delve into the LLM's robustness to varying ASR performance conditions and scrutinize the aspects of in-context learning which prove the most influential.
%
\end{abstract}
\section{Introduction}
\label{sec:intro}


In spoken language understanding (SLU) systems, typically a spoken utterance is first transcribed by automated speech recognition (ASR) to yield a hypothesis, called the ``$1$-best", before being fed into a more conventional NLU pipeline. The $1$-best often has errors that negatively impact downstream performance. 

To mitigate this issue, there has been research on using additional information from ASR. Relevant strategies include feeding n-best alternative hypotheses to a transformer language model~\cite{ganesan-etal-2021-n}, extending RNN language models to lattices~\cite{ladhak2016lattice}, flattening lattices and masking attention between parallel lattice paths~\cite{huang2020adapting}, and using word confusion networks (WCNs) to compute weighted sums of word options or paths~\cite{8462030, liu2020jointly}. 


The idea to supply additional information from ASR to decoder-based LLMs~\cite{le2023voicebox, wang2023neural} for SLU is under-explored, as previous work has focused on encoder models. One recent work addressed this by providing noisy in-context examples from an ASR model, with optional chain-of-thought (CoT) error correction for intent classification~\cite{he2023chatgpt}. In the CoT setup, authors give ASR transcripts of commands from the SLURP benchmark~\cite{bastianelli2020slurp}, with descriptions of ASR errors found in the in-context examples (e.g. ```Stop price' doesn't make sense, which could be a mistranscription of `Stock price'..."), instructing the model to correct such errors in test examples before determining the intent. The performance with and without CoT error correction lagged behind use of the ground-truth text. While results improved after incorporating CoT error correction into the prompt, only the $1$-best transcript is provided for test examples, lacking any specific ambiguities or word options. Furthermore, the range of potential ASR error types broadly found in test data is unlikely to be adequately represented through a few in-context examples.


In this paper, we propose a simple representation of WCNs which can be fed to off-the-shelf LLMs for downstream SLU tasks. Specifically, we study whether through in-context learning~\cite{brown2020language} and our representations, LLMs can achieve improved robustness to ASR errors and ambiguities compared to using the $1$-best transcript. Our findings indicate that model size is critical, and that in-context instruction and examples each provide improved performance.

    

\begin{figure}[ht!]
    \centering
    \includegraphics[width=0.50\textwidth]{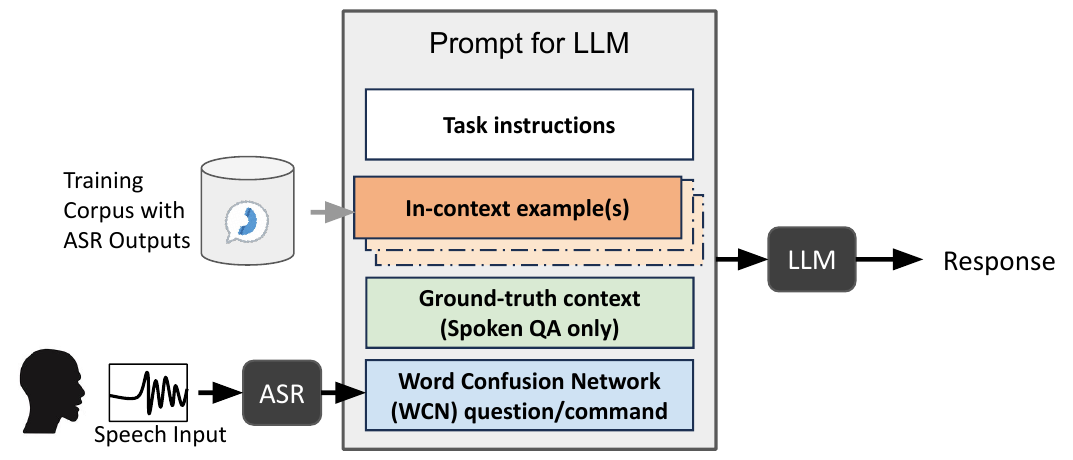}
    \caption{Proposed in-context learning framework using a WCN question transcript for Spoken Language Understanding (SLU).}
    \label{fig:1:framework}
\end{figure}


    \begin{figure}[htb]
    \begin{minipage}[b]{0.97\linewidth}
      \centering
      \centerline{\includegraphics[width=9cm]{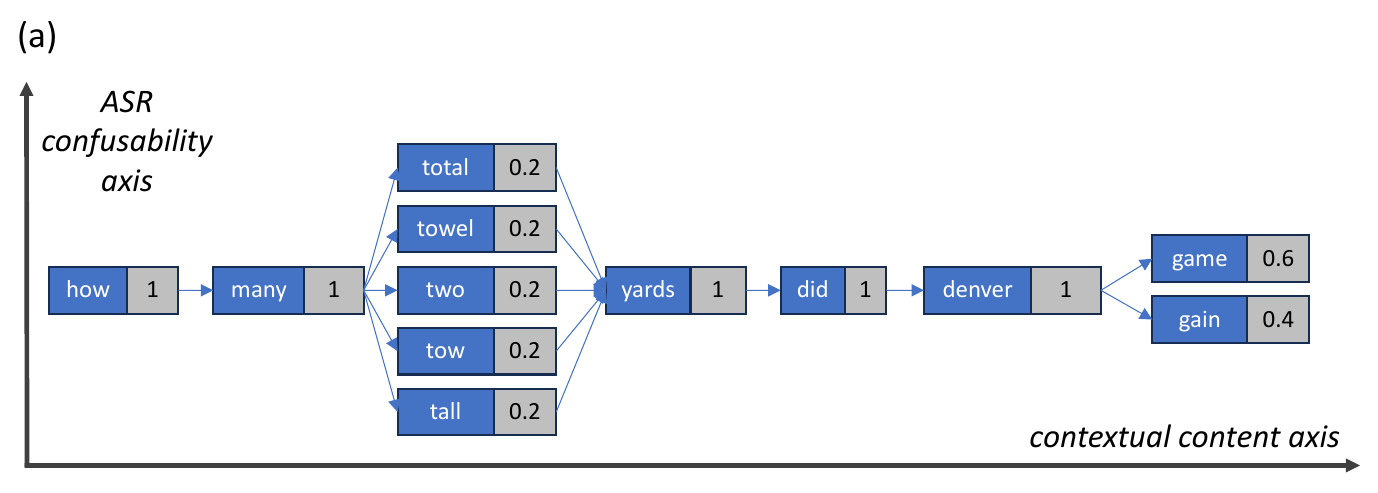}}
      \vspace{0.2cm}
     \end{minipage}
    \begin{minipage}[b]{1.0\linewidth}
        \centering
        (b) \texttt{how many towel|two|total|tow|tall yards did denver game|gain?}
      \vspace{0.2cm}
    \end{minipage}
    \begin{minipage}[b]{1.0\linewidth}
        \centering
        (c) \texttt{how many total yards did denver game|gain?}
    \end{minipage}
    \caption{Constructing LLM input from raw WCN: (a) Raw WCN, (b) LLM input from WCN, and (c) LLM input from WCN after filtering based on word posteriors ($p \geq 0.3$).}
    \label{fig:2:conf_net}
    \end{figure}



\section{Methods}
\label{sec:methods}

As shown in the pipeline depicted in Figure~\ref{fig:1:framework}, we use an ASR model to extract a WCN from speech, and insert a representation of this WCN into a prompt template before passing it to an LLM.

\subsection{Word confusion networks (WCN) input representation}
\label{ssec:conf}

\subsubsection{ASR system and WCN construction}
\label{sssec:asr}

We use an off-the-shelf Emformer-RNNT ASR model,%
\footnote{https://pytorch.org/audio/main/generated/torchaudio.models.emformer\\\_rnnt\_model} pretrained on the LibriSpeech dataset~\cite{panayotov2015librispeech}
to transcribe spoken data and extract lattices, which can be used to create WCNs using the Kaldi toolkit~\cite{Povey_ASRU2011, XU2011802}. As shown in the example in Fig. \ref{fig:2:conf_net}(a), WCNs are derived from lattices and contain ambiguous word options aligned along the \textit{contextual content axis}. Each word option is accompanied by a posterior, which sum to one over all alternative hypotheses at a given position in the utterance. We use WCNs rather than lattices because they contain more paths, while lending themselves to a simpler, more compact string representation.

\subsubsection{LLM inputs from WCNs}
\label{sssec:llm_conf_input}

Starting with a WCN for each example, we create an input that can be interpreted by an LLM by flattening the ASR confusability axis. We separate alternative words with a separator symbol, either a slash ``$/$" or pipe ``$|$", chosen due to their function in common written text to represent alternative options. Optionally, we can also filter options based on a posterior threshold. This process of constructing an input from a WCN is demonstrated in Fig.~\ref{fig:2:conf_net}(a-c). Our method requires minimal preprocessing and only modifications to the input itself.  We note that a limitation of this format is that it cannot represent ambiguity between word hypotheses and word deletions (null words), which can also result from ASR output. For the present study we simply delete null alternatives.

\subsection{In-context learning for SLU with WCNs}
\label{ssec:icl_slu}

Since our separators may be represented in different contexts within the training data,
we optionally include a short description of the in-context significance of the WCN separator symbol (``$/$" or ``$|$") in our prompt (known as ``WCN instruction"). In one-shot experiments, WCN instruction is always included, and in-context examples use the same transcript source as the test example, i.e., we use the $1$-best transcript for the in-context example when testing with a $1$-best transcript, a WCN representation when testing on a WCN representation, etc. Thus, the in-context example provides not only a task demonstration, but also an example of ASR output representations.


\subsection{Language models}
\label{ssec:models}

For our experiments, we use the \textit{BLOOMZ-560M} and \textit{BLOOMZ-3B} models~\cite{muennighoff2022crosslingual}, which were fine-tuned on a multilingual multi-task dataset. We also use a fixed \textit{GPT-3.5-turbo} model~\cite{brown2020language}. 

\textbf{BLOOMZ} ($560$M \& $3$B): BLOOM stands as the inaugural open-source LLM trained on a French government-sponsored supercomputer~\cite{scao2022bloom}. Its successor, BLOOMZ, undergoes continuous fine-tuning using multi-task incremental learning~\cite{wang2022super} and generative-aware prompting paradigms~\cite{muennighoff2022crosslingual}. The foundational training of the model uses an extensive public repository of HuggingFace datasets~\cite{lhoest2021datasets}, which aggregates up to $1.61$ TB of text across $46$ natural languages and $13$ programming languages. Notably, our evaluation datasets are exempted from this collection.

\textbf{GPT-3.5} (approx. $175$B): Building upon the foundational GPT-3 model~\cite{brown2020language, ouyangtraining}, GPT-3.5 employs the principles of reinforcement learning from human feedback (RLHF)~\cite{christiano2017deep} to further its training. This paradigm culminates in \textit{GPT-3.5-turbo}, showcasing enhanced zero-shot learning capabilities.
It's important to clarify that in our investigations, we have opted for a \textbf{consistent version} of GPT-3.5, rather than directly interfacing with various versions of ChatGPT.

We opt to use each LLM off-the-shelf, rather than fine-tuning them with ASR transcripts or our WCN transcript notation. Hence, the efficacy of our methods hinges on the models' ability to recall from its training data that our chosen separators (``$/$" and ``$|$") indicate word choices, as well as the capacity of the models to learn in-context.

\subsection{Baselines}
\label{ssec:baselines}

To evaluate the efficacy of our WCN representation, we compare results to the following baseline setups, with varying LLM inputs:

\textbf{Ground-truth transcripts:} the ground-truth transcript provides an upper bound, as performance is naturally expected to degrade when using any ASR output.

\textbf{ASR \textit{n}-best oracle:} the hypothesis from the $n$-best list with the lowest edit distance to the ground-truth transcript (using the ASR score as a tie-breaker). This provides a more realistic upper bound, since the WCN input is unlikely to provide more useful information than the most correct $n$-best hypothesis.

\textbf{ASR $\textbf{1}$-best:} the simplest approach when using ASR output, this provides a baseline for comparison, with the goal to improve over \textbf{ASR $\textbf{1}$-best} by enriching the LLM input with WCNs.

\section{A Study of Spoken Question Answering}
\label{sec:nmsqa}

\subsection{NMSQA dataset and task}
\label{ssec:nmsqa_data}

The Natural Multi-speaker Spoken Question Answering (NMSQA) dataset~\cite{lin22c_interspeech} contains spoken versions of examples from the SQuAD dataset~\cite{rajpurkar2016squad, lee2018spoken}. The train and dev splits were created using a text-to-speech (TTS) system, while crowdworkers were hired to compile the spoken test set. As with SQuAD, the examples in the NMSQA dataset consist of a paragraph of context, a question, and a corresponding answer which can be found directly in the context. Due to computing limitations and small size of the original test set, we partition the original NMSQA dev set into a small development set consisting of the final $1000$ examples, for tuning the posterior threshold, and a test set consisting of the first $4,800$ examples. Details on the NMSQA dataset (and our dev/test splits) are shown in Table~\ref{table:nmsqa_details}.

\begin{table}
\begin{center}
\caption{NMSQA statistics}
\label{table:nmsqa_details}
\begin{tabular}{ c|c|c c }
\textbf{Split} & \# questions & $1$-best WER\% & $n$-best oracle WER\% \\
\hline
train & $87125$ & $21.52\%$ & $14.73\%$ \\
dev & $1000$ & $22.09\%$ & $15.43\%$ \\
test & $4800$ & $19.35\%$ & $12.82\%$ \\
\end{tabular}
\end{center}
\end{table}

Both SQuAD and NMSQA were created with \textit{extractive} question answering in mind, in which a model is tasked with identifying the span in the context that contains the answer. Under our setup, we prompt the LLM to provide the answer given a ground-truth text context and transcribed question, and measure performance using unigram F1 and exact match (EM) metrics.

\subsection{Prompt design and spoken QA results}
\label{ssec:nmsqa_results}

The LLM prompt~\cite{wu2022promptchainer, chang2023speechprompt, yang2021voice2series, wu2023speechgen, chang2022speechprompt} includes a short description of the SQA task, followed by the context and question transcript. We envision a scenario in which an agent has access to a context document, and a user asks questions that can be answered from the context. Accordingly, the ground-truth context is always used, while the question uses the various transcript sources previously described (with optional WCN instruction, as indicated in Section \ref{ssec:icl_slu}).

\begin{table}
\begin{center}
\caption{NMSQA test-set zero-shot results.}
\label{table:nmsqa_zeroshot_results}
\adjustbox{width=0.48\textwidth}{
\begin{tabular}{ r r | c c | c c | c c }
 & & & & & & & \\
\textbf{Transcript} & \textbf{WCN} & \multicolumn{2}{|c|}{\textbf{\textit{BLOOMZ-560M}}} & \multicolumn{2}{|c|}{\textbf{\textit{BLOOMZ-3B}}} & \multicolumn{2}{|c}{\textbf{\textit{GPT-3.5}}} \\
\textbf{source} & \textbf{instr.} & F1 & EM & F1 & EM & F1 & EM \\
\hline
\textit{Ground-truth} & & $\textit{58.9\%}$ & $\textit{37.0\%}$ & $\textit{67.1\%}$ & $\textit{35.6\%}$ & $\textit{71.7\%}$ & $\textit{42.5\%}$ \\
\hline
$1$-best & & $\textbf{48.3\%}$ & $\textbf{29.9\%}$ & $\textbf{58.4\%}$ & $\textbf{38.2\%}$ & $\textbf{64.5\%}$ & $\textbf{36.5\%}$ \\
``$/$" & & $44.0\%$ & $27.0\%$ & $57.2\%$ & $37.1\%$ & $50.5\%$ & $21.0\%$ \\
``$/$" & \checkmark & $44.4\%$ & $26.8\%$ & $54.3\%$ & $30.2\%$ & $53.6\%$ & $25.5\%$ \\
``$/$" ($p \geq 0.3$) & & $47.6\%$ & $29.3\%$ & $58.0\%$ & $37.7\%$ & $56.1\%$ & $27.7\%$ \\
``$/$" ($p \geq 0.3$) & \checkmark & $47.0\%$ & $28.6\%$ & $56.7\%$ & $31.6\%$ & $61.9\%$ & $33.7\%$ \\
``$|$" & & $42.1\%$ & $26.2\%$ & $57.1\%$ & $37.1\%$ & $53.0\%$ & $24.6\%$ \\
``$|$" & \checkmark & $43.5\%$ & $26.6\%$ & $53.6\%$ & $31.5\%$ & $57.4\%$ & $31.0\%$ \\
``$|$" ($p \geq 0.3$) & & $47.6\%$ & $29.3\%$ & $58.0\%$ & $37.7\%$ & $56.1\%$ & $27.3\%$ \\
``$|$" ($p \geq 0.3$) & \checkmark & $46.4\%$ & $28.5\%$ & $56.4\%$ & $31.0\%$ & $63.2\%$ & $35.3\%$ \\
\end{tabular}}
\end{center}
\end{table}


 Based on development set performance, we use a posterior threshold of $0.3$. Zero-shot results with optional WCN instruction are reported in Table~\ref{table:nmsqa_zeroshot_results}. It is immediately evident that with each model, none of the WCN representations achieve better performance than the $1$-best. In addition, the \textit{BLOOMZ} models show limited ability to learn via WCN instruction, with results generally degrading when using WCN input with posterior filtering. However, use of the WCN instruction in the prompt led to significantly improved results for \textit{GPT-3.5}, with relative improvements ranging from $5\%$ to $29\%$ over prompting without WCN instruction.

\begin{table}
\begin{center}
\caption{NMSQA test-set one-shot results..}
\label{table:nmsqa_oneshot_results}
\adjustbox{scale=1.0}{
\begin{tabular}{ r | c c | c c }
 & \multicolumn{2}{c|}{\textbf{\textit{BLOOMZ-3B}}} & \multicolumn{2}{c}{\textbf{\textit{GPT-3.5}}} \\
\textbf{Transcript source} & F1 & EM & F1 & EM \\
\hline
\textit{Ground-truth} &  $\textit{58.1\%}$ & $\textit{36.7\%}$ & $\textit{76.2\%}$ & $\textit{47.9\%}$ \\
\textit{$n$-best oracle} &  \cellcolor[HTML]{EFEFEF}$\textit{47.7\%}$ & \cellcolor[HTML]{EFEFEF}$\textit{29.7\%}$ & \cellcolor[HTML]{EFEFEF}$\textit{72.4\%}$ & \cellcolor[HTML]{EFEFEF}$\textit{45.1\%}$ \\
\hline
$1$-best & $\textbf{43.6}\%$ & $\textbf{27.1}\%$ & $70.2\%$ & $43.0\%$ \\
``$/$" & $31.6\%$ & $18.7\%$ & $70.4\%$ & $43.8\%$ \\
``$/$" ($p \geq 0.3$) & $38.0\%$ & $23.9\%$ & $70.3\%$ & $44.0\%$ \\
``$|$" & $32.1\%$ & $19.1\%$ & $69.6\%$ & $44.1\%$ \\
``$|$" ($p \geq 0.3$) & $38.6\%$ & $24.3\%$ & $\textbf{70.5}\%$ & $\textbf{44.7}\%$ 
\end{tabular}}
\end{center}
\end{table}

One-shot performance with WCN instruction is reported in Table~\ref{table:nmsqa_oneshot_results}.
We use one randomly-chosen \texttt{(context, question, answer)} example from the training split, with the same transcript source for the in-context example and the test question. The \textit{BLOOMZ-560M} model displayed poor ability to learn from in-context examples, and is thus omitted. While $1$-best transcripts still produce the best results for \textit{BLOOMZ-3B}, the WCN inputs produce similar or better results with \textit{GPT-3.5}. Specifically, using ``$|$" with posterior filtering improved between $0\%$ and $4\%$ (relative) over the $1$-best result. For the EM metric, the best WCN method closes $81\%$ of the performance gap between the $1$-best baseline and the $n$-best oracle upper bound.

\section{A Study of Intent Classification}
\label{sec:atis}

\subsection{ATIS dataset and task}
\label{ssec:atis_data}

The Airline Travel Information Systems (ATIS) corpus~\cite{hemphill-etal-1990-atis} contains spoken requests that were designed to be answered via a corresponding database, and has been widely used in spoken intent classification (IC) and slot filling (SF) tasks. The label space for the IC task consists of seventeen individual intents and is highly skewed, with the top seven classes 
combined covering $94\%$ of the entire dataset. The test set contains $809$ command/intent pairs, with a $1$-best WER of $8.40\%$ and $n$-best oracle WER of $5.31\%$.


\subsection{Prompt design and results}
\label{ssec:atis_methods}

Similar to the NMSQA task, the LLM prompt for ATIS consists of a short description of the task with a list of the intent classes in alphabetical order, followed by the user command transcript. 

\begin{table}
\begin{center}
\caption{ATIS zero-shot intent classification results (accuracy).}
\adjustbox{scale=0.9}{
\label{table:atis_results}
\begin{tabular}{ r | c | c | c }
\textbf{Transcript source} & \textbf{\textit{BLOOMZ-560M}} & \textbf{\textit{BLOOMZ-3B}} & \textbf{\textit{GPT-3.5}} \\
\hline
Ground-truth & $\textit{69.5\%}$ & $\textit{73.6\%}$ & $\textit{88.1\%}$ \\
$n$-best oracle & \cellcolor[HTML]{EFEFEF}$\textit{67.5\%}$ & \cellcolor[HTML]{EFEFEF}$\textit{73.4\%}$ & \cellcolor[HTML]{EFEFEF}$\textit{86.0\%}$ \\
\hline
$1$-best & $\textbf{66.8\%}$ & $73.6\%$ & $84.6\%$ \\
``$/$" & $51.9\%$ & $73.6\%$ & $84.8\%$ \\
``$/$" ($p \geq 0.3$) & $65.4\%$ & $\textbf{73.7\%}$ & $84.8\%$ \\
``$|$" & $61.3\%$ & $73.3\%$ & $84.8\%$ \\
``$|$" ($p \geq 0.3$) & $64.3\%$ & $73.6\%$ & $\textbf{85.0\%}$ \\
\end{tabular}}
\end{center}
\end{table}

We report the accuracy for each model using each transcript source in Table \ref{table:atis_results}. For the smallest model, using $1$-best transcripts resulted in the best performance, with WCN performance lagging by $1.5-15\%$. In addition, posterior filtering was essential for achieving results that approach the $1$-best performance. Meanwhile, the \textit{GPT-3.5} model achieved the best performance using the WCN with ``$|$" and posterior filtering, with all WCN setups producing slightly better performance than the $1$-best. Considering the $n$-best oracle as an upper bound, the best WCN setup is able to close $33\%$ of the gap between the $1$-best baseline and this upper bound.





\section{Analysis and Discussion}
\label{sec:analysis}

Based on the findings in Sections~\ref{ssec:nmsqa_results} and~\ref{ssec:atis_methods}, the following trends emerge that predict under what conditions WCN representations may be able to achieve comparable or better performance than $1$-best transcripts:
\begin{itemize}
    \item \textbf{Instruction:} using WCN instructions improves performance with \textit{GPT-3.5}.
    \item \textbf{In-context examples:} adding a single in-context example improves performance with \textit{GPT-3.5}.
    \item \textbf{Posterior filtering:} especially in the noisier NMSQA setting, filtering WCNs by posteriors achieves the best performance.
    \item \textbf{Model size:} the large \textit{GPT-3.5} model is able to achieve WCN performance exceeding that from the $1$-best. Given that the key difference between \textit{GPT-3.5} and the \textit{BLOOMZ} models is size (in addition to RLHF training), we hypothesize that the ability to disambiguate between confused word options could be an emergent property of larger models.
\end{itemize}

\subsection{Impact of ASR errors and failure analysis}
\label{ssec:asr_impact}

To assess the impact of ASR errors on the performance of our methods, we divide the NMSQA dev set into two partitions: examples where the $1$-best $\text{WER\%}=0\%$ or $\text{WER\%}\neq0\%$. The EM results when using $1$-best transcripts and WCN inputs are reported in Table~\ref{tab:asr_impact}, with WCN transcripts showing over $4\%$ relative improvement over using the $1$-best when ASR errors exist. Meanwhile, improvements are more modest in the no-error scenario.

\begin{table}[]
    \centering
    \caption{NMSQA test-set one-shot EM in examples with and without ASR errors (overall WER$=22\%$).}
    \begin{adjustbox}{width=0.48\textwidth}
    \begin{tabular}{c c | c c c}
        Subset & $\#$ examples & $1$-best & ``$|$" ($p \geq 0.3$) & Rel. Improv \\
        \hline
        w/out ASR errors & $548$ & $48.2\%$ & $48.6\%$ & $0.83\%$ \\
        w/ ASR errors & $4252$ & $42.3\%$ & $44.1\%$ & $4.26\%$ \\
    \end{tabular}
    \end{adjustbox}
    \label{tab:asr_impact}
\end{table}


\begin{figure}[ht!]
    \centering
    \includegraphics[width=0.5\textwidth]{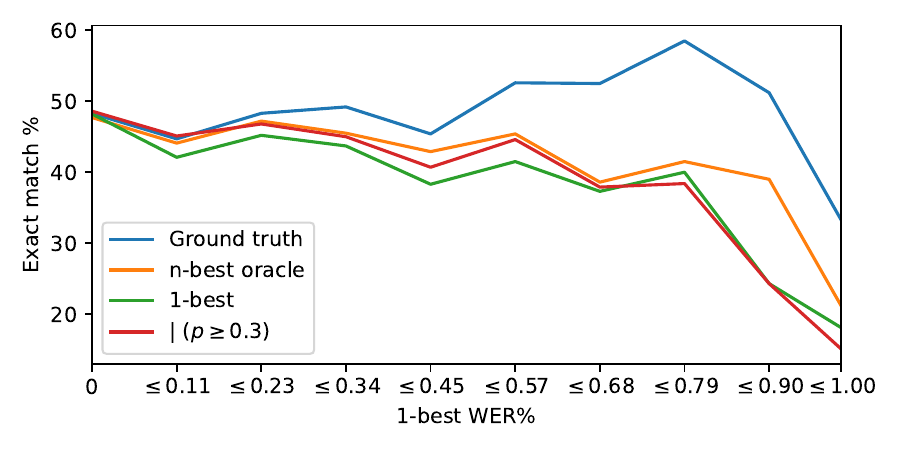}
    \caption{NMSQA test-set one-shot EM by question WER bin.}
    \label{fig:3:em_wer}
\end{figure}

We more closely examine performance across a range of ASR conditions in Figure~\ref{fig:3:em_wer}. By sorting NMSQA test examples into bins based on their question WER\%, it is shown that while performance is similar among examples with no word errors (echoing Table~\ref{tab:asr_impact}), the WCN representation performance exceeds that of $1$-best transcripts over a wide range of WER\% scenarios. However, it is important to note that under very noisy conditions (over $\sim70\%$ WER), using the $1$-best is preferable to WCNs.

To illustrate the trends in Figure~\ref{fig:3:em_wer}, selected examples are displayed in Table~\ref{table:examples}, for \textbf{(a)} zero, \textbf{(b)} medium, and \textbf{(c)} high WER\% scenarios. While our WCN methods are comparable to using the $1$-best (see Table~\ref{tab:asr_impact}) in $0\%$ WER examples, \textbf{(a)} contains an example in which using the WCN adds unnecessary word options and even removes relevant words which were present in the $1$-best, resulting the incorrect answer. Conversely, the WCN question in \textbf{(b)} recovers relevant words that were deleted in the $1$-best, and the LLM generates the correct response. Finally, in the high WER\% example shown in \textbf{(c)}, the WCN input introduces noise with incorrect options for several words. To this point, we note that with posterior filtering ($p \geq 0.3$), WCN questions in the $0\%$ WER bin contain an average of $1.04$ options per word, with this metric steadily increasing to $1.12$ options/word in the highest WER bin, in which transcripts can become saturated with noisy (and often incorrect) options.

\begin{table}
\begin{center}
\caption{NMSQA test-set examples with varying $1$-best WER\%.}
\label{table:examples}
\adjustbox{scale=0.85}{
\begin{tabular}{ c | c }

\multicolumn{1}{c|}{} & \multicolumn{1}{c}{} \\
\multicolumn{1}{c|}{\multirow{-2}{*}{\textbf{Transcript source}}}   & \multicolumn{1}{c}{\multirow{-2}{*}{\textbf{Question? \textcolor{gray}{Generated answer} (\textcolor[HTML]{027148}{correct}/\textcolor{red}{incorrect})}}}    \\
\hline
\multicolumn{2}{l}{\textbf{(a)} Zero WER\%} \\
\hline
\multicolumn{1}{c|}{Ground-truth} & what do these teachers not do? \textcolor[HTML]{027148}{teach by rote} \\ 
\hline
\multicolumn{1}{c|}{$1$-best} & what do these teachers not do? \textcolor[HTML]{027148}{teach by rote} \\ 
\hline
\multicolumn{1}{c|}{\multirow{2}{6em}{``$|$" ($p \geq 0.3$)}} & what do these teachers knocked$|$knock \\ 
 & to$|$due? \textcolor{red}{knock} \\ 
\hline
\multicolumn{2}{l}{\textbf{(b)} Medium WER\%} \\
\hline
\multicolumn{1}{c|}{\multirow{2}{6em}{Ground-truth}} & when was his article published \\ 
 & in century magazine? \textcolor[HTML]{027148}{nineteen hundred} \\ 
\hline
\multirow{2}{6em}{$1$-best} & one was his article published in century \\
 & magaz? \textcolor{red}{the problem of increasing human energy} \\ 
\hline
\multirow{2}{6em}{``$|$" ($p \geq 0.3$)} & one$|$when was his article published in \\
 & century$|$sanctary magazine$|$magaz? \textcolor[HTML]{027148}{nineteen hundred} \\ 
\hline
\multicolumn{2}{l}{\textbf{(c)} High WER\%} \\
\hline
\multirow{2}{6em}{Ground-truth} & who did denver beat in the  \\ 
 & afc championship? \textcolor[HTML]{027148}{the new england patriots} \\ 
\hline
\multirow{2}{6em}{$1$-best} & who did them for beating the \\
 & a f sea champions? \textcolor[HTML]{027148}{the new england patriots} \\ 
\hline
\multirow{3}{6em}{``$|$" ($p \geq 0.3$)} & who did them for beating in the \\
 & a$|$aye f i see$|$sea champions$|$championship? \\
 & \textcolor{red}{the arizona cardinals} \\ 
\end{tabular}}
\end{center}
\end{table}



\subsection{In-context example transcript format}

In one-shot NMSQA experiments, the in-context example that we provide has the same transcript representation as the test example. Thus, the one-shot setup potentially helps via two avenues:  (a) demonstration of the SQA task and (b) demonstration of the ASR or WCN representation. To assess the impact of each, we evaluate the NMSQA test set with WCN inputs (using ``$|$" and $p \geq 0.3$), using an in-context example with ground-truth transcripts rather than the corresponding WCN transcripts, effectively removing the impact of (b) above. As shown in Table~\ref{tab:IC_impact}, the performance difference between using ground-truth transcripts versus WCN transcripts in the in-context examples is minimal, indicating that most of the improvement when adding an in-context example is due to demonstration of the task, rather than demonstration of the WCN representation.

\begin{table}[]
    \centering
    \caption{NMSQA test-set one-shot results with ``$|$" ($p \geq 0.3$) transcripts and varying in-context transcript type.}
    \begin{adjustbox}{width=0.4\textwidth}
    \begin{tabular}{c | c c}
        \textbf{In-context transcript type} & \textbf{F1} & \textbf{EM} \\
        \hline
        N/A (zero-shot) & $63.2\%$ & $35.3\%$ \\
        \hline
        Ground-truth & $70.2\%$ &  $\textbf{44.8}\%$ \\
        ``$|$" ($p \geq 0.3$) & $\textbf{70.5}\%$ &  $44.7\%$ \\
    \end{tabular}
    \end{adjustbox}
    \label{tab:IC_impact}
\end{table}






\subsection{Current limitations and future work}

We consider simplicity and lack of a need for model fine-tuning to be the key advantages of our approach, showing that current off-the-shelf LLMs can exhibit improved robustness to ASR transcripts by representing ambiguity in the input. However, this success was limited to \textit{GPT-3.5}, and was not mirrored by the smaller models. Since the use of WCN yielded worse results with \textit{BLOOMZ} models, the ability of smaller models to ingest and understand WCN representations remains an open issue. It is likely that fine-tuning the LLMs directly on ASR transcripts and WCN representations would give better results in future studies.



\section{Conclusions}
\label{sec:conclusions}

We have proposed a simple method for representing ambiguities from ASR output for use in SLU tasks using LLMs. Our strategy requires only preprocessing the input, and yields improved results over 1-best transcripts when using larger models.

\vfill\pagebreak



\clearpage
\small
\bibliographystyle{IEEEbib}

\end{document}